\documentclass{article}
\usepackage{spconf,graphicx}
\usepackage{comment}
\usepackage{booktabs}
\usepackage{ragged2e} 
\usepackage{makecell, multirow, tabularx}
\usepackage{amsmath}
\usepackage{amsfonts} 

\title{T-Pixel2Mesh: Combining Global and Local Transformer for 3D Mesh Generation from a Single Image}
\name{Shijie Zhang$^{1}$, Boyan Jiang$^{1}$, Keke He$^{2}$, Junwei Zhu$^{2}$, Ying Tai$^{3}$, Chengjie Wang$^{2}$, Yinda Zhang$^{4}$, Yanwei Fu$^{1\star}$}
\address{$^{1}$ School of Data Science, Fudan University, Shanghai, China \\
$^{2}$Youtu Lab, Tencent, Shanghai, China\\
$^{3}$School of Intelligence Science and Technology, Nanjing University, Suzhou, China\\
$^{4}$Google, USA\\
}

\begin{document}
%
\maketitle
\begin{abstract}
Pixel2Mesh (P2M) is a classical approach for reconstructing 3D shapes from a single color image through coarse-to-fine mesh deformation. Although P2M is capable of generating plausible global shapes, its Graph Convolution Network (GCN) often produces overly smooth results, causing the loss of fine-grained geometry details. Moreover, P2M generates non-credible features for occluded regions and struggles with the domain gap from synthetic data to real-world images, which is a common challenge for single-view 3D reconstruction methods. To address these challenges, we propose a novel Transformer-boosted architecture, named T-Pixel2Mesh, inspired by the coarse-to-fine approach of P2M. 
Specifically, we use a global Transformer to control the holistic shape and a local Transformer to progressively refine the local geometry details with graph-based point upsampling. To enhance real-world reconstruction, we present the simple yet effective Linear Scale Search (LSS), which serves as prompt tuning during the input preprocessing. Our experiments on ShapeNet demonstrate state-of-the-art performance, while results on real-world data show the generalization capability.
\end{abstract}
\section{Introduction}
\label{sec:intro}

Generating accurate 3D shapes from few 2D observations, \emph{e.g.} merely a single color image, is a challenging yet essential task for simulating human 3D visual perception. Deep neural networks have led to several single- and multi-view shape generation approaches \cite{Occupancy_Networks,QiLWSG18,wen2022pixel2mesh++}. While these methods often produce plausible holistic shapes, capturing detailed local geometry and generalizing to real-world images remains difficult. The main challenge lies in effectively leveraging limited 2D visual cues and integrating them with underlying representations for 3D shape generation.
To address this challenge, tremendous effort has been devoted to various learning-based 3D representation methods, including 3D voxels \cite{choy20163d,wang2017cnn}, implicit surface representation \cite{Occupancy_Networks,xu2019disn,li2021d2im}, point clouds \cite{fan2017point,QiLWSG18,wen20223d} and meshes \cite{pixel2mesh,wen2022pixel2mesh++}. However, voxels demand high memory footprint cubed with grid resolution, implicit surface representation methods require a time-consuming surface extraction to obtain final mesh, while light-weight point clouds is not desirable for many downstream applications such as graphics rendering and shape manipulation. In contrast, mesh representation supports fast inference and local adjustments, thus is more suitable for a wide range of applications.
In this paper, we propose a novel framework, T-Pixel2mesh, for single-view 3D mesh generation based on explicit mesh representation. 
Our model builds on the previous Pixel2Mesh (P2M) \cite{pixel2mesh}, enhancing the mesh deformation process by a hybrid attention mechanism, Transformer-based Deformation Module (TDM) in a coarse-to-fine manner. Our TDM leverages global and local self-attention mechanisms to capture holistic shapes while refining detailed geometry.
The key insight is that global attention could automatically learn attention maps to filter out less meaningful features from the occluded regions, while local attention could refine the detailed geometry with the spatial information from neighbors. 

Furthermore, to enhance the generalization capabilities for real-world reconstruction, we present a simple yet effective Linear Scale Search (LSS) approach. Generalization capability is crucial for learning-based reconstruction methods, considering the difficulty or even impossibility of obtaining 3D ground truths for certain real-world images. Experiments show that LSS approach, which can be intuitively viewed as prompt tuning \cite{lester2021power} during preprocessing, improves performance on real-world images with varying camera intrinsics and camera-to-object distances. 

\begin{figure*}
\begin{centering}
\begin{tabular}{cc}
\begin{tabular}{c}
\includegraphics[width=0.5\linewidth]{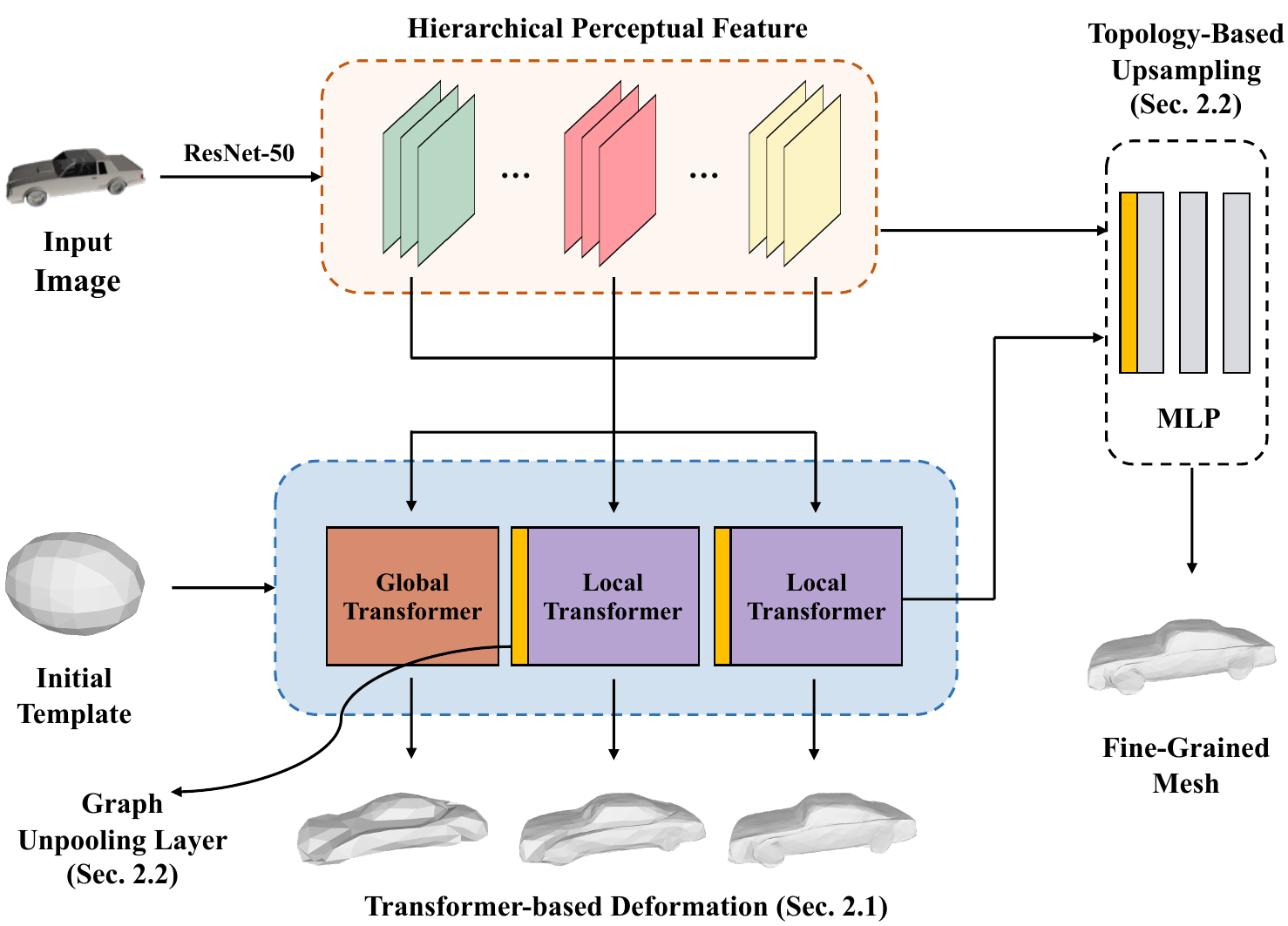}\tabularnewline
\end{tabular} & %
\begin{tabular}{c}
 \includegraphics[width=0.45\linewidth]{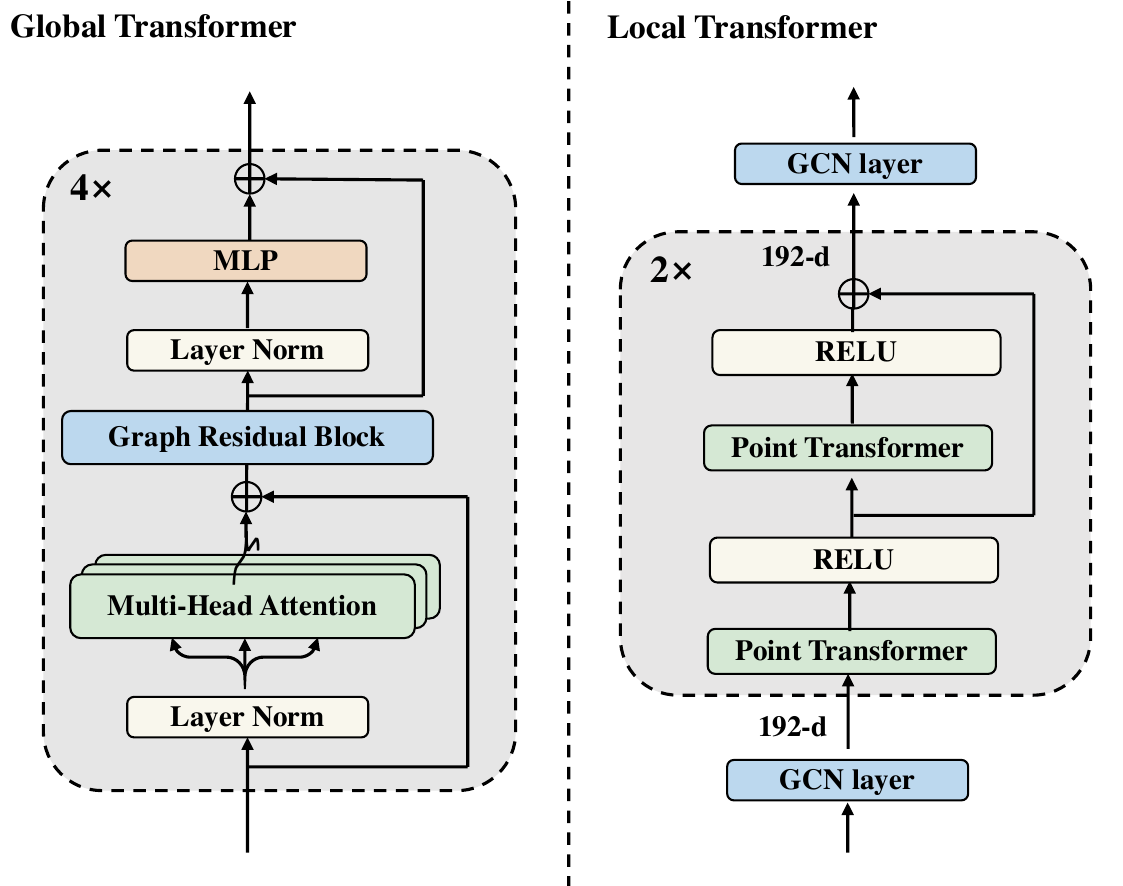}\tabularnewline
\end{tabular}\tabularnewline
(a) The overview of our T-Pixel2Mesh.  & (b) Architectures of Transformers.\tabularnewline
\end{tabular}
\par\end{centering}
\vspace{-0.1in}
\caption{(a) The pipeline of T-Pixel2Mesh. (b) Architecture details of Global Transformer Block and Local Transformer Block. \label{fig:overview}}
\vspace{-0.15in}
\end{figure*}

Our contributions are summarized as follows:
1) A novel network T-Pixel2Mesh, the first approach to combine global and local Transformer for mesh generation, with its global attention map for better shape in occluded parts and local Transformer for finer local geometry details.
2) A LSS approach to improve the robustness of reconstruction from in-the-wild images.
3) Comprehensive experiments on synthesized and real datasets demonstrate superior performance.

\section{Proposed Method}
\label{sec:method}
\vspace{-0.1in}
\noindent \textbf{Overview}
Given a single color image as input, T-P2M directly generates the corresponding 3D mesh model, as visualized in Figure~\ref{fig:overview}(a). We first utilize ResNet-50 \cite{he2016deep} pretrained on ImageNet to extract perceptual features from the input image, creating hierarchical feature maps for multi-scale feature pooling. This process captures local geometry and high-level semantic information from 4 layers of feature maps. Vertex features are acquired by projecting onto hierarchical feature maps, pooling pixel-aligned vectors, and concatenating them with coordinates, resulting in a 3843-dimensional vector.

Starting with an initial ellipsoid template, our Transformer-based Deformation Module (TDM) progressively deforms it towards the target shape, as described in Sec. \ref{sec:tdm}. TDM consists of three deformation blocks. The first block features a global Transformer that performs self-attention operations between all mesh vertices, while the last two blocks contain local Transformers that consider features from neighboring vertices in 3D space. To enhance the representation of high-frequency geometric details, we increase the vertex count using graph-based upsampling (Sec. \ref{sec:final_unsmp}) after each deformation block. 
In Sec. \ref{sec:lss}, we discuss the Linear Scale Search (LSS) approach designed for real-world images.

\begin{figure}[h]
 \centering
 \vspace{-0.25in}
 \includegraphics[width=1\linewidth]{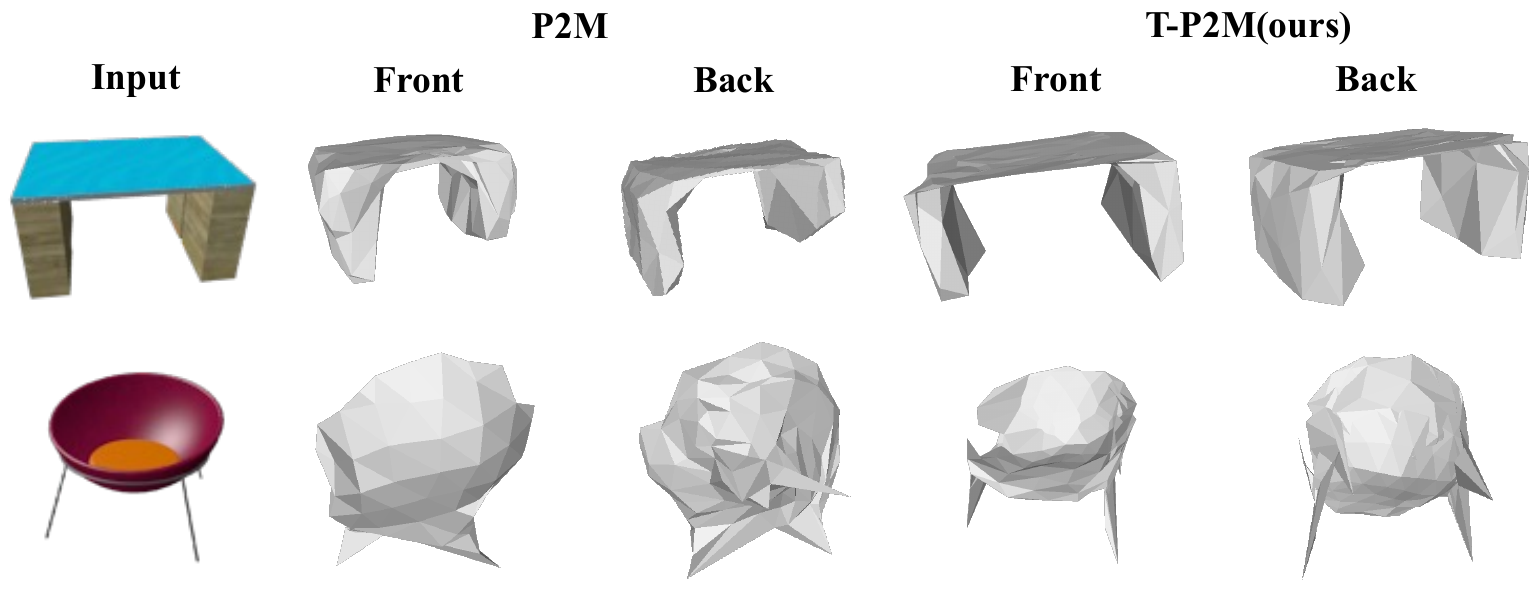}
 \vspace{-0.35in}
 \caption{Coarse shape reconstruction results (meshes with 156 vertices). With the help of global Transformer, our T-P2M not only recovers more accurate holistic shapes in occluded areas (`back' column) but also achieves better symmetry.\label{fig:global-trans}}
  \vspace{-0.3in}
\end{figure}
\subsection{Transformer-based Deformation Module}\label{sec:tdm}
\noindent \textbf{Global Transformer}
In the first mesh deformation stage, we aim to build a coarse holistic shape with relatively small number of points, namely $156$ vertices of the initial ellipsoid mesh. 
Our global Transformer is a generalized version of vanilla Transformer \cite{vaswani2017attention} in dealing with meshes derived from Mesh Graphormer \cite{graphormer}.
Specifically, we first project the vertices of initial template to the hierarchical feature maps to pool pixel-aligned feature vectors. We also take the intermediate output of $7\times7$ resolution as Global Feature Vectors and concatenate the 49 vectors with the 156 vertex vectors.
We then treat all $205$ vertices as a token sequence $X$, to perform self-attention for exchanging geometric information. In other words, this process automatically learns attention maps to emphasize informative features while filtering out less useful ones.
\begin{figure*}
 \centering 
 \includegraphics[width=0.95 \linewidth]{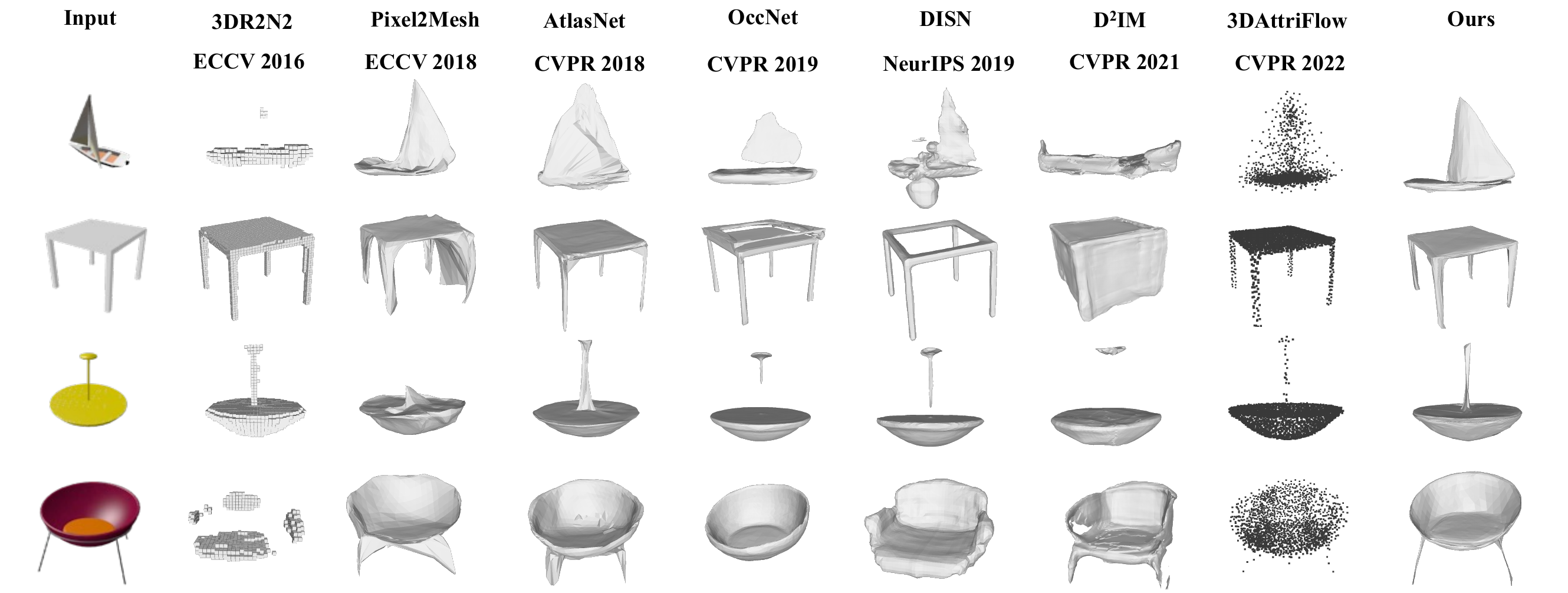} 
 \vspace{-0.15in}
 \caption{The qualitative results of single-view mesh generation task on ShapeNet dataset.
 \label{fig:single-recons}}
  \vspace{-0.2in}
\end{figure*}

Formally, our Transformer encoder consumes $X$ with Multi-Head Self-Attention (MHSA) layer:
\begin{equation}
    {z_i}^h = \mathrm{softmax}(\frac{(Q^h {K^h}^T)}{\sqrt{d}})V^h.
    \label{eq:scalar trans}
\end{equation}
with queries $Q=X{W_Q}$, keys $K=X{W_K}$, values $V=X{W_V}$ and head number $h$.
The output of MHSA is processed by a Multi-Layer Perceptron (MLP) with residual connection. The detailed architecture is shown in Fig. \ref{fig:overview}(b). 
We further add a Graph Residual Block before the MLP to augment the global features with local information from connected neighbors since we empirically find it boosts performance as shown in Sec.\ref{sec:ablation}. The graph convolutional layer \cite{gcn_layer} works as:
\begin{equation}
    {x_i}^{n+1} = w_0 {x_i}^n + \sum_{j \in N(i)} w_1 {x_j}^n,
\end{equation}
where ${x_i}^{n} \in \mathbb{R}^{d_n}$ and ${x_i}^{n+1} \in \mathbb{R}^{d_{n+1}}$ are the tokens on vertex $i$ before and after convolution, $N(i)$ denotes the neighbouring vertices of $x_i$,  $w_0$ and $w_1$ represent learnable matrices. 
Fig.\ref{fig:global-trans} shows the benefits of our Global Transformer Block by comparing reconstruction results after the first block in P2M (a GCN block) and in our T-P2M.


\noindent \textbf{Local Transformer} 
Global Transformer provides a coarse shape estimation but loses fine-grained geometry details. While graph unpooling in P2M could increase point number, global attention between all vertices is computationally expensive. The limited receptive field of P2M's GCN hinders local deformation capture, resulting in overly smooth outputs.

Instead, inspired by \cite{point_transformer}, we use local Transformers in the last two deformation blocks of TDM, aggregating geometry information from $k$-nearest neighbors. 
Local Transformer, different from scalar attention \cite{vaswani2017attention} as in Eq. \ref{eq:scalar trans}, uses vector attention \cite{vector_trans} since it requires fewer computational resources and has scalability to more points \cite{point_transformer,xiang2021snowflakenet}, formulated as: 
\begin{equation}
    z_i=\sum_{x_j \in \chi(i) } \mathrm{softmax} \left( \gamma \left( \varphi (x_i) - \phi(x_j) +\delta  \right) \right) \odot \left( \alpha(x_i) +\delta \right)
\end{equation}
$\chi(i)$ denotes the $k$ nearest points of $x_i$, $\varphi$, $\phi$ and $\alpha$ are linear projection layers, $\gamma$ is an MLP to produce attention vectors and $\odot$ denotes Hadamard product. 
$\delta = \theta (c_i - c_j)$ is a positional embedding getting information from relative point position
where $c_i$ and $c_j$ are vertex coordinates and $\theta$ is an MLP with SiLU. 
It is an adaption of \cite{vector_trans} to adjust attention weight for each feature channel, namely each vertex, instead of one-for-all. 
Our local Transformer enables each vertex to gather features from any assigned $k$-nearest neighbor vertices, thus is convenient to scale up to higher resolutions. 
We set $k=16$ in second deformation stage of 618 vertices and $k=64$ in third stage of 2466 vertices. Design details are in Fig. \ref{fig:overview}(b).
\vspace{-0.1in}
\subsection{Graph-based Point Upsampling}\label{sec:final_unsmp}
To support the coarse-to-fine deformation strategy, we use an upsampling layer before each point Transformer block to increase the point count. Our point upsampling layer shares operations with the edge-based graph unpooling layer used in P2M but has different design motivations. We upsample point clouds based on the underlying graph for better structure preservation and more representative added points.
Technically, for each face on mesh, we subdivide its edges by adding a vertex at the midpoint and connecting this new vertex with the two end-point of the edge.
Though TDM generating plausible results, the final output with 2466 vertices struggles to handle high-frequency details, often producing artifacts around thin structures such as chair legs.
Instead of an additional local Transformer Block, we construct an efficient final upsampling layer that combines graph-based upsampling with a small MLP, since we empirically find it sufficient to capture fine-grained geometry yet avoid extra computational costs. We leverage the intermediate features from the last block and the pixel-aligned features as input of our final unsampling layer, and it directly predicts the 3D coordinates for 9858 vertices. We validate its effectiveness in Sec. \ref{sec:ablation}.
\vspace{-0.1in}
\subsection{Linear Scale Search} \label{sec:lss}
It's challenging to reconstruct 3D meshes from casually captured single-view images using a model trained merely on synthetic ShapeNet dataset due to the domain gap, as in-the-wild images are captured with various devices and under complex conditions.
To address this, we preprocess input to closely resemble training data by controlling the object's scale in the image.
We first use a background mask from \cite{kirillov2023segment} to crop a bounding image of the object, pad it to a square size, and control the object's scale by border padding. The padded border width $\mathbf{p}=s \times h$, where $s$ and $h$ represent the scale coefficient and squared bounding image size. We choose the best value of $s$ based on the final mesh quality, linearly searching within the range of $\left[0.2, 0.4\right]$.

\begin{table*}
\centering
\resizebox{\textwidth}{!}{
    \begin{tabular}{l|c|ccccccccccccc}
    \toprule
    Methods &Mean  &Plane    &Bench  &Cabinet   &Car   &Chair   &Display    &Lamp    &Loud.   &Rifle &Sofa  &Table  &Tele. &Vessel   \\ \midrule
    3DR2N2 \cite{choy20163d}   &5.41   &4.94   &4.80  &4.25  &4.73    &5.75   &5.85    &10.64    &5.96    &4.02    &4.72  &5.29   &4.37   &5.07   \\
    Pixel2Mesh  \cite{pixel2mesh}   &5.27  &5.36 &5.14    &4.85    &4.69    &5.77    &5.28    &6.87    &6.17    &4.21   &5.34   &5.13   &4.22   &5.48   \\
    AtlasNet \cite{groueix2018atlasnet}  &3.59 &2.60    &3.20    &3.66    &3.07    &4.09    &4.16    &4.98    &4.91  &2.20  &3.80   &3.36   &3.20   &3.40   \\
    OccNet \cite{Occupancy_Networks}   &4.15   &3.19   &3.31    &3.54    &3.69    &4.08    &4.84    &7.55    &5.47    &2.97  &3.97   &3.74   &3.16   &4.43   \\
    DISN \cite{xu2019disn}  & 3.96 & 3.98    & 3.51 & 4.12   & 3.08   & 3.26  & 4.62  & 7.70  & 6.47  & 1.99  & 3.66  & 3.16  & 2.82  & 3.12    \\
    $\rm D^2$IM \cite{li2021d2im}  & 3.73 & 3.58    & 3.12 & 3.85   & 3.48   & 3.29  & 4.22  & 5.57  & 5.61  & 2.44  & 3.91  & 3.56  & \textbf{2.45}  & 3.39    \\
    3DAttriFlow \cite{wen20223d} & 3.02    & \textbf{2.11} & \textbf{2.71}   & 2.66   & \textbf{2.50}  & 3.33  & 3.60  & 4.55  & 4.16  & \textbf{1.94}  & 3.24  & 2.85  & 2.66  & \textbf{2.96}   \\
    \midrule
    T-P2M (Ours) &\textbf{2.96}  & 2.95  &2.85  &\textbf{2.59}  &2.66  & \textbf{3.18} &\textbf{3.40} &\textbf{3.68}  &\textbf{3.56}  &2.37  &\textbf{3.19}  &\textbf{2.39}  &2.50  &3.20  \\
    \bottomrule
    \end{tabular}
}
\vspace{-0.15in}
\caption{Quantitative comparisons of single-view 3D shape generation on ShapeNet  (Chamfer Distance-$L1$, lower is better).}
\label{tab:single-recons}
\vspace{-0.1in}
\end{table*}

\section{EXPERIMENTAL RESULTS AND ANALYSIS}
\noindent \textbf{Dataset and Metric}
Following the previous single-view 3D reconstruction methods \cite{pixel2mesh,Occupancy_Networks,groueix2018atlasnet}, we use 13 common object categories (Tab. \ref{tab:single-recons}) from ShapeNet. Then we use in-the-wild images from Pix3D \cite{sun2018pix3d} and CO3D \cite{reizenstein2021common} datasets to evaluate the performance on real-world data.
We use Chamfer Distance-$L1$ metric for quantitative comparison.

\noindent \textbf{Implementation Details}
Our framework receives input images of size 224$\times$224. 
We use Adam optimizer with batch size of 48, initial learning rate of $5e^{-4}$, dropping to $1.5e^{-5}$ after 30 epochs and weight decay of $1e^{-6}$. We train all the modules in our model end-to-end for 50 epochs. The total training process consumes 3 days on 4 V100 GPUs.

\noindent \textbf{Single-View Shape Generation}
We evaluate the performance of our method on the task of single-view 3D mesh generation by comparing with 3DR2N2 \cite{choy20163d}, Pixel2Mesh \cite{pixel2mesh}, AtlasNet \cite{groueix2018atlasnet}, OccNet \cite{Occupancy_Networks}, DISN \cite{xu2019disn}, $\rm D^{2}$IM \cite{li2021d2im} and 3DAttriFlow \cite{wen20223d}.
The quantitative comparisons on Chamfer-Distance are shown in Tab. \ref{tab:single-recons}. The numbers are borrowed from \cite{wen20223d,li2021d2im}. Our T-P2M clearly outperforms all the baseline methods on average score and most of the 13 categories from ShapeNet.
We also provide some qualitative results in Fig. \ref{fig:single-recons}, which shows 3DR2N2 is limited by grid resolution, Pixel2Mesh produces artifacts such as cracks on the surface and faces between table legs, AtlasNet has unsmooth parts and unsatisfactory thin structures, OccNet and DISN suffers from floating parts and longer marching-cube operation time, D2IM underperforms in categories like guns and cars and 3DAttriFlow produces a sparse point cloud lacking surface information, making further rendering or topology-based refinement inconvenient. In contrast, T-P2M excels in presenting complete geometry and fine details.
\begin{figure}
 \centering 
 \includegraphics[width=1\linewidth]{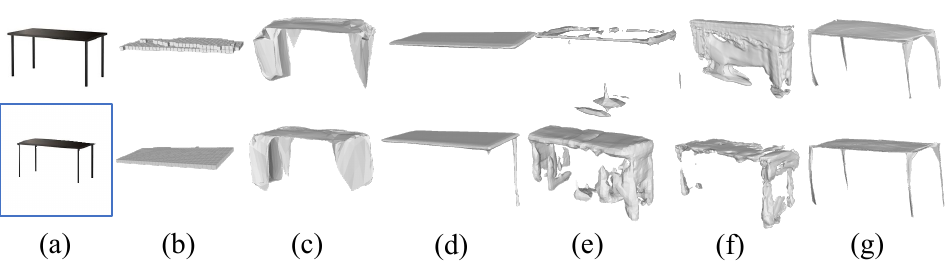} 
   \vspace{-0.25in}
 \caption{Mesh generation on in-the-wild images, upper row without LSS and lower row with LSS. (a) Input; (b) 3DR2N2; (c) Pixel2Mesh; (d) OccNet; (e) DISN; (f) $\rm D^2$IM; (g) Ours.}
 \label{fig:real-world}
  \vspace{-0.15in}
\end{figure}
\noindent \textbf{Mesh Generation in the Wild}
We choose some real-world images from Pix3D \cite{sun2018pix3d} and CO3D \cite{reizenstein2021common} as input to perform single-view mesh generation for evaluation. We use LSS for preprocessing to choose the best reconstruction as described in Sec. \ref{sec:lss}.
In Fig. \ref{fig:real-world}, all the baseline methods have a significant performance drop comparing with the generation quality on ShapeNet, since the input image is in the data domain far from training data.
When equipped with LSS, our method can capture most details in the input images and produce the best reconstructions, thanks to the strong capacity provided by our hybrid attention mechanism, which effectively learns correlations between spatial vertices and facilitates the deformation process.
The performance of other methods also improves with LSS, which may offer insights into addressing the reconstruction problem in complex real scenarios.

\begin{table}
    \small
    \centering
  \scalebox{0.8}{
    \begin{tabular}{c|c|c|c|c}
        \toprule 
         & Stage-1 & Stage-2 & Stage-3 & Chamfer-L1 $\downarrow$ \tabularnewline
        \midrule 
        \multirow{3}{*}{w/o up.} 
        & Local & Local & Local & 3.646\tabularnewline
         & Global & Local & Local & 3.623\tabularnewline
        \midrule 
       
        & GCN & GCN & GCN & 3.088 \tabularnewline
    w/ up.     & Local & Local & Local & 3.019 \tabularnewline
        & Global(w/o GRB) & Local & Local & 3.032 \tabularnewline
      (Ours)  & Global & Local & Local & \textbf{2.973} \tabularnewline
        \bottomrule 
    \end{tabular}}
\caption{Ablation study.}
\vspace{-0.15in}
\label{tab:module-ablation}
\end{table}

\noindent \textbf{Ablation Study} \label{sec:ablation}
We evaluate our T-P2M framework's major modules, including including global Transformer (Global), local Transformer (Local), graph convolutional network (GCN) and Graph Upsampling (up.) in Tab. \ref{tab:module-ablation}. Both attention mechanism and Graph Upsampling boost performance, but using only Local Transformer worsens results due to missing holistic shape control. An all GCN-based framework sets off the effect of our attention-based design by contrast. Row 5, without GRB, demonstrates GRB's enhancement of neighboring features in Global Transformer Block. Overall, combining global and local Transformers in mesh deformation and using the final upsampling module yields the best performance, validating our design choices.

\section{Conclusion}

We present a Transformer-boosted framework for 3D mesh generation from single-view images, which combines global and local Transformer to exchange geometry information between vertices. Such hybrid attention mechanism effectively manages holistic 3D shapes while capturing local details. Our method is the first to combine global and local Transformer for mesh generation.
Experiments demonstrate our superior performance on both synthesis and real-world data.

\vfill\pagebreak

\bibliographystyle{IEEEbib}
\bibliography{sample-base}

\end{document}